# A METHODOLOGY FOR EXPLAINABLE LARGE LANGUAGE MODELS WITH INTEGRATED GRADIENTS AND LINGUISTIC ANALYSIS IN TEXT CLASSIFICATION

Marina Ribeiro*[1,2], Bárbara Malcorra*[2], Natália B. Mota[2,3], Rodrigo Wilkens[4,5], Aline Villavicencio[5,6], Lilian C. Hubner[7], César Rennó-Costa[#1]

[1] Bioinformatics Multidisciplinary Environment (BioME), Digital Metropolis Institute (IMD), Federal University of Rio Grande do Norte (UFRN), Natal (RN), Brazil

[2] Research Department at Mobile Brain, Mobile Brain, Rio de Janeiro (RJ), Brazil

[3] Institute of Psychiatry (IPUB), Federal University of Rio de Janeiro (UFRJ), Rio de Janeiro (RJ), Brazil

[4] Department of Computer Science, The University of Exeter, Exeter, UK

[5] Institute for Data Science and Artificial Intelligence at the University of Exeter, Exeter, UK

[6] Department of Computer Science, The University of Sheffield, Sheffield, UK

[7] School of Humanities, Pontifical Catholic University of Rio Grande do Sul (PUCRS), Porto Alegre (RS), Brazil

* Equal author contribution

# Corresponding author: cesar@imd.ufrn.br

## ABSTRACT

Neurological disorders that affect speech production, such as Alzheimer's Disease (AD), significantly impact the lives of both patients and caregivers, whether through social, psycho-emotional effects or other aspects not yet fully understood. Recent advancements in Large Language Model (LLM) architectures have developed many tools to identify representative features of neurological disorders through spontaneous speech. However, LLMs typically lack interpretability, meaning they do not provide clear and specific reasons for their decisions. Therefore, there is a need for methods capable of identifying the representative features of neurological disorders in speech and explaining clearly why these features are relevant. This paper presents an explainable LLM method, named SLIME (Statistical and Linguistic Insights for Model Explanation), capable of identifying lexical components representative of AD and indicating which components are most important for the LLM's decision. In developing this method, we used an English-language dataset consisting of transcriptions from the Cookie Theft picture description task. The LLM Bidirectional Encoder Representations from Transformers (BERT) classified the textual descriptions as either AD or control groups. To identify representative lexical features and determine which are most relevant to the model's decision, we used a pipeline involving Integrated Gradients (IG), Linguistic Inquiry and Word Count (LIWC), and statistical analysis. Our method demonstrates that BERT leverages lexical components that reflect a reduction in social references in AD and identifies which further improve the LLM's accuracy. Thus, we provide an explainability tool that improves confidence in applying LLMs to neurological clinical contexts, particularly in the study of neurodegeneration.

**Introduction**

Identifying mental disorders is a complex task requiring the administration of combined neuropsychological and cognitive assessment, which, compared to other medical conditions, relies on a higher degree of subjectiveness in the assessments. This limitation has spurred the development of automated tools, which are becoming increasingly valuable in the field (Epelbaum & Cacciamani, 2023). Recent advances in computer science, particularly in Natural Language Processing (NLP) and automated speech analysis, have proven to be a promising tool in identifying and predicting psychiatric conditions (Elvevåg et al., 2007; Bedi et al., 2015; Mota et al., 2017). For instance, automated speech analysis has been successfully used to detect subtle mental state changes in emergent psychosis, with methods like Latent Semantic Analysis - LSA (Landauer & Dumais, 1997), a technique that analyzes relationships between a set of documents and the terms they contain by producing a lower-dimensional representation of words and documents based on their co-occurrence patterns, and *SpeechGraphs* (Mota et al., 2012), a method that uses graph theory to analyze the organization of speech by examining word recurrence patterns, providing diagnostic accuracy comparable to clinical ratings (Mota et al., 2017). The most recent automated speech analysis approaches typically rely on Large Language Models (LLMs), which are artificial intelligence (AI) systems trained on extensive text data that extract both topological information and linguistic features from the text, often in combination with machine learning models (De La Fuente Garcia et al., 2020). These tools offer standardized approaches for supporting the detection of psychiatric disorders, creating opportunities for early diagnosis and intervention that could significantly advance the field.

Despite the remarkable accuracy of automated speech analysis methods, the current SOTA relies on "black box" AI models that offer limited or no transparency regarding the decision-making process (Doshi-Velez & Kim, 2017; Adadi & Berrada, 2018). For instance, various studies employing models such as BERT - Bidirectional Encoder Representations from Transformers - (Devlin et al., 2019), a type of LLM, have demonstrated high performance in tasks like detecting Alzheimer's disease (AD) on a text response (Balagopalan et al., 2021; Guo et al., 2021; Yuan et al., 2021; Zhu et al., 2021). Yet, these models base their decision on a linear mapping of a multidimensional feature space that is not interpretable for a human actor (Bibal et al., 2022). Other modeling approaches, such as graph-based models, also suffer from similar opacity as their decision-making process is based on graph attributes such as the largest connected component and the number of repeated edges (Mota et al., 2012, 2014, 2023). Reduced transparency presents significant challenges for researchers, clinicians, and individuals seeking diagnostic insights, as the inability to interpret or trust the model's outcomes can limit their application in clinical settings (Amann et al., 2020; Abdullah et al., 2021; Loch et al., 2022).

Explainability has emerged as a critical feature of AI systems to address these concerns. Explainability refers to the AI system's ability to elucidate the reasoning behind its decisions or predictions, fostering stakeholder trust and understanding (Gunning et al., 2019; Vilone & Longo, 2021; Ali et al., 2023). It is particularly crucial in healthcare, where AI-driven decisions can have profound implications, and understanding the rationale behind these outputs is essential. An example of an explainability technique is AI-driven clinical image processing, such as X-ray analysis, where heatmaps are overlaid on the original image to highlight the regions most relevant to the model's decision. It allows clinicians to visually interpret which areas influence the AI's classification (Selvaraju et al., 2020; Preechakul et al., 2022). However, achieving similar levels of explainability in text analysis presents unique challenges. Unlike images, where spatial relationships and features can be visually highlighted, textual data involves complex linguistic structures and semantic relationships that are less straightforward to map. Providing clear and interpretable explanations for decisions made by text-based models, particularly those as intricate as BERT or graph-based models, remains a significant challenge in NLP.

This paper presents a novel methodology for lexical explainability in text classification tasks called SLIME (Statistical and Linguistic Insights for Model Explanation). SLIME assigns a relevance score, attribution, to each input feature, indicating how much it influences the model's output. The current implementation utilizes the Integrated Gradients (IG) method (Sundararajan et al., 2017). Like heatmaps highlight essential areas in image processing, the attributions in SLIME identify which words or word components significantly impact the model's decision. Building on these attributions, we introduce a method to identify statistically significant linguistic attributes of these words, revealing the most relevant terms and providing insights into their critical linguistic features, allowing for explanatory annotation at a linguistic level. To demonstrate our approach, we use labels from the Linguistic Inquiry and Word Count (LIWC) toolkit (Boyd et al., 2022). We validate the method using a dataset of audio recordings and transcripts from the Cookie Theft picture description task (Goodglass & Kaplan, 1972), achieving an accuracy of 87% in a 5-fold cross-validation experiment. We also discuss the practical application of our methodology in both scientific research and clinical settings, highlighting its potential to enhance the transparency and trustworthiness of AI in bringing complementary evidence for clinical diagnosis. We acknowledge the limitations and challenges of pursuing lexical explainability in text classification tasks.

**Methods**

*Speech dataset*

The dataset used for training in the classification task was developed for the ADReSS challenge (Luz et al., 2020). The challenge aimed to detect Alzheimer's disease (AD) through various speech analysis tasks. The dataset was designed to have an equal number of subjects in both the AD and control groups, balanced across age and sex (Table 1). The challenge consisted of two tasks: classifying Alzheimer's through developing speech analysis algorithms to identify the condition and predicting the cognitive decline score based on the standard Mini-Mental State Examination (MMSE) test (Folstein et al., 1975). The dataset includes audio recordings of descriptive narratives of the Cookie Theft picture description task, their transcriptions, the subject's age and sex, and their MMSE scores. However, the MMSE score was not used in our work. The Cookie Theft picture description task (Goodglass & Kaplan, 1972) is a standard test for identifying aphasia. Still, it has been used as a complementary assessment tool for various clinical conditions, including mental disorders (Berube et al., 2022; Li et al., 2019) and neurodegenerative diseases (Boschi, 2017; Cummings, 2019; Lai & Lin, 2012; March et al., 2006; Mueller et al., 2018). It presents an image depicting a woman washing dishes while the sink is overflowing, with two children attempting to reach a cookie jar on the upper counter by standing on a stool. Essentially, the subjects need to describe the presented image. Therefore, the data used to train, validate, and test the classification model consists of 156 descriptive narratives (78 AD, 78 control) of the Cookie Theft picture description task.

Table 1: Dataset description

| Group | N | Age | Sex (f \| m) | MMSE |
|---|---|---|---|---|
| **AD** | 78 | 66.6 ± 6.8 | 43 \| 35 | 17.8 ± 5.5 |
| **Control** | 78 | 66.3 ± 6.6 | 43 \| 35 | 28.6 ± 3.5 |

Legend: N = number of participants; AD = Alzheimer's disease; f = female / m = male; MMSE = Mini-Mental State Examination.

*LLM model for text classification*

Pre-trained neural language models are commonly used for specific tasks where computational power is limited, or the dataset is small. Utilizing a pre-trained model allows us to leverage its existing knowledge and merge it with the new dataset's knowledge. We should employ a pre-trained model with 156 descriptive reports in our dataset. Retraining the model on our dataset makes it an expert for the classification task. The process of imparting this specialized knowledge to the model is termed fine-tuning.

We use the pre-trained model BERT (Devlin et al., 2019) based on the Transformer architecture (Vaswani et al., 2017). This model utilizes a self-attention mechanism to map input and output comprehensively. Self-attention provides positional representations for each token in the input text, enabling its relevance to be influenced by its context or position. BERT's advantage lies in its ability to apply self-attention in a bidirectional manner, as it was initially trained to predict a specific part of the text that could occur before or after any other text. Therefore, BERT is an advantageous model for textual applications where context is crucial. Additionally, BERT has been trained in various languages, including Brazilian Portuguese (Souza et al., 2020), making validating our method in other languages easier.

Fine-tuning for the classification task involves adding a linear layer to the original architecture. The model's input is text, with a maximum sentence size of 512 tokens. The output from the linear layer provides a classification score between 0 and 1, indicating the classes of interest. In this case, we train the model on the labeled ADReSS dataset, where 0 represents the control, and 1 represents the AD condition. We employ 5-fold cross-validation, which entails dividing the dataset into five smaller sets, training on four parts, and leaving one for validation. This process is repeated five times. Accuracy is calculated for each fold, and the model selected for our explainability pipeline corresponds to the fold with the highest validation accuracy, 87%. In detail, we trained the BERT base cased model using the AdamW optimizer (Loshchilov & Hutter, 2017), with learning rate = 2e-5 and Adam's hyperparameter eps = 1e-8, for 50 epochs with a batch size of 1. The loss function used was Binary Cross-Entropy with Logits.

*Model explainability with Integrated Gradients (IG)*

To explain our model, we want to identify which specific elements (or tokens) in the picture description led the model to classify a subject as either a control or an Alzheimer's patient.

Integrated Gradients (Sundararajan et al., 2017) involves a neural network model $F : \mathfrak{R} \to [0, 1]$ that takes an input $x = (x_1, ..., x_n) \in \mathfrak{R}$ and produces a prediction $F(x)$. The IG scores, denoted as $IG(x) = (a_1, ..., a_n) \in \mathfrak{R}$, rate the contribution of each element $x_i$ in the input toward the prediction. A baseline $x'$ is defined for the IG computation to obtain meaningful attributions, a reference for what to ignore $x$. In NLP tasks, the baseline typically consists of the embedding values representing the tokens' absence. IG is a reliable approach for model explainability, as it is based on the mathematical axioms of Sensitivity and Implementation Invariance. Sensitivity refers to how the model $F$ responds to changes in its inner parameters, such as weights and bias. Therefore, if $F$ returns null gradients or asymptotic values for different predictions, this indicates low sensitivity. The use of a baseline ensures that IG satisfies sensitivity. Implementation invariance means that models with different architectures should produce the same output given the same input. This implies that the "path" between input and output, i.e., the model's hidden layers, should not affect the attribution computation.

These axioms can be mathematically expressed in terms of the gradient path integral:

$$F(x) - F(x') = \int_{x'}^{x} \nabla_\gamma F \cdot d\gamma, \qquad \text{(Eq. 1)}$$

which represents the gathering of the significant contributions through the path $\gamma$, from the baseline $x'$ to the input $x$, to make the prediction $F(x)$.

Considering the path $\gamma(\alpha) = x' + \alpha(x - x')$ parametrized by the contrast adjustment $\alpha \in [0, 1]$, the IG is defined as:

$$IG_i(x, x') := (x - x') \int_0^1 \frac{\partial F(x' + \alpha(x - x'))}{\partial x_i} d\alpha. \qquad \text{(Eq. 2)}$$

Here, $F$ is the customized BERT for classification, $x_i$ is the picture description's embedded tokens, $F(x)$ is the probability classification score, and $IG_i$ accounts for the relevance of each embedded token $x_i$ in discriminating between control and Alzheimer's subjects. We calculate the IG in BERT's embedding layer according to Kokhlikyan et al. (2020).

*Definition of Linguistic Features - Linguistic Inquiry and Word Count (LIWC)*

LIWC (Boyd et al., 2022) is a text categorization software that counts words from a validated dictionary. It takes plain text or a table as input (specifying the column containing the text to be analyzed). It produces a table showing the proportions of category words in the input text. We utilized the 2022 English dictionary, consisting of 117 linguistic categories. For our token-level analysis, we excluded 6 categories that rely on the overall text context: word count (*WC*), analytic thinking metric (*Analytic*), leadership language (*clout*), perceived honesty (*Authentic*), degree of positivity or negativity (*Tone*), and words per sentence (*WPS*). In our study, the input text is a token, and its LIWC category value is either 0 or 100, depending on its absence or presence in the category. For example, if "she" is a token of interest, running LIWC would show that this token is present in the following categories: *Dic, Linguistic, function, pronoun, ppron, shehe, Social, socrefs, and female*. Therefore, a single token can belong to multiple categories. It is important to note that some categories have subcategories, such as *ppron* (personal pronouns), which is a subcategory of *pronoun*. However, our analysis treated each category independently and did not consider these subdivisions.

*Identification of Relevant Linguistic Features - Statistical Analysis*

We developed a statistical test to verify whether a specific linguistic feature of LIWC is a relevant explainability concept. Our statistical analysis yields two key insights: first, it identifies which linguistic features are most relevant for each group, and second, it assesses their impact on classification. Figure 1 presents a schematic representation of the explainability method.

First, we compare the distribution densities of the attributions for a given linguistic feature with those of all tokens to determine if the feature is more relevant for one group (Figure 2A). Values less than or equal to zero are associated with the control group, while values higher than zero are associated with the AD group. Then, we conduct 5000 random subsamples of the same size as the linguistic feature of interest and check if it is above the 95th percentile or below the 5th percentile of the subsample distribution. If either condition is met, the feature significantly contributes to one of the groups. For example, in Figure 2B, the feature *female* (normative references to female figures such as she, her, girl, woman) contributes to the decision of the control group. After that, we isolate the feature of interest to estimate its classification power between the groups using AUC (Area Under the Curve) analysis (Figure 2C), which is a method for evaluating classification model performance using ROC (Receiver Operating Characteristic) curves. The AUC represents the model's ability to distinguish between positive and negative cases, with 1 indicating perfect prediction and 0.5 indicating random guessing. Last, the AUC value of the feature is then compared to the mean AUC values of the random subsamples (Figure 2D). If the feature's AUC is above the 95th percentile or below the 5th percentile of the subsample distribution, we conclude there is a significant impact on AUC, positive or negative, respectively. The difference between the feature's AUC and the mean of the distribution provides an essential variable for visualizing the results of the explainability method.

*Method validation*

To validate our method, we present a basic test distinguishing between groups using only LIWC without employing a neural language model for classification. From the outset, it is essential to emphasize that SLIME is not intended to be the most accurate or discriminative classifier. Instead, it serves as an explainability method for large language models, whose rationale we present in this article.

The test presented here involves providing the full descriptions from the Cookie Theft picture task to the LIWC text annotation software. The output provided by LIWC is a proportion (mean relative to the total word count of the text) indicating how much a given linguistic category is present in the input text. For the comparative analysis of linguistic category discrimination by group (AD and control), we used the two-sided Mann-Whitney U test with an AUC measure for comparison with SLIME's results. For validation analysis, AUC is calculated as the ratio between the Mann-Whitney U statistic for the control sample and the product of the sample sizes (78 for each group). The AUC value for the AD sample is calculated as one minus the AUC for the control sample. The p-value for statistical significance is given by

Bonferroni correction (alpha/number of categories analyzed), with alpha = 0.05 and 111 categories analyzed (p = 0.0004). Figure 6 presents the validation analysis.

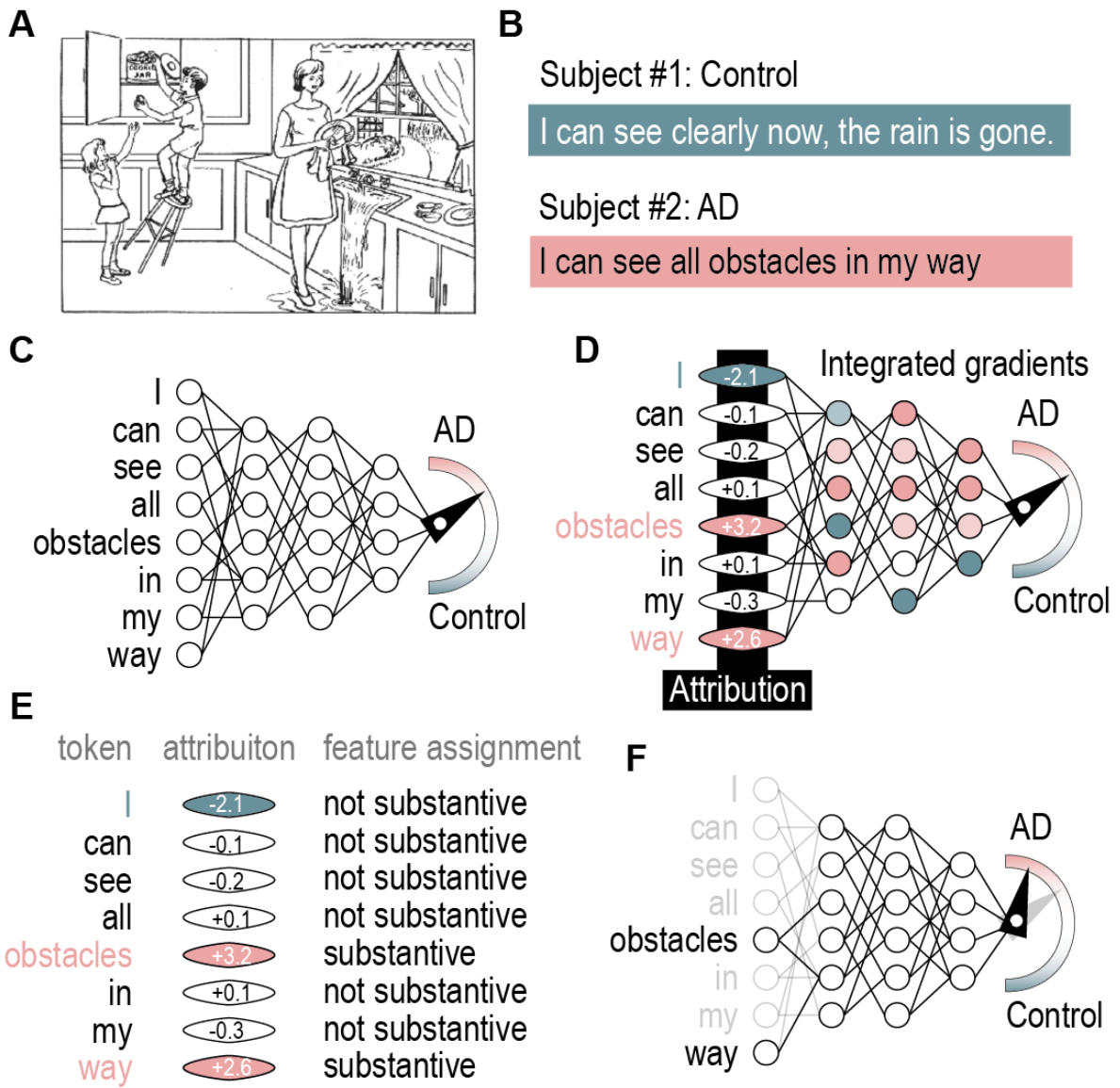

**Figure 1: The explainability method. A.** The method involves utilizing an oral and spontaneous description of an image (e.g., the Cookie Theft picture description task) as its basis. **B.** Descriptions from the control and AD groups are utilized for a classification model. **C.** A pre-trained BERT model is employed and subsequently fine-tuned with the collected descriptions. **D.** Integrated gradients are utilized to gain access to the final layer of the model, providing numerical attributions at the token level to indicate the relevance of each token in the decision for group classification. **E.** Each token can be tagged for characterization using LIWC for linguistic tagging. **F.**

Through statistical analysis, it is possible to infer the most relevant attributes for classification and their linguistic characteristics.

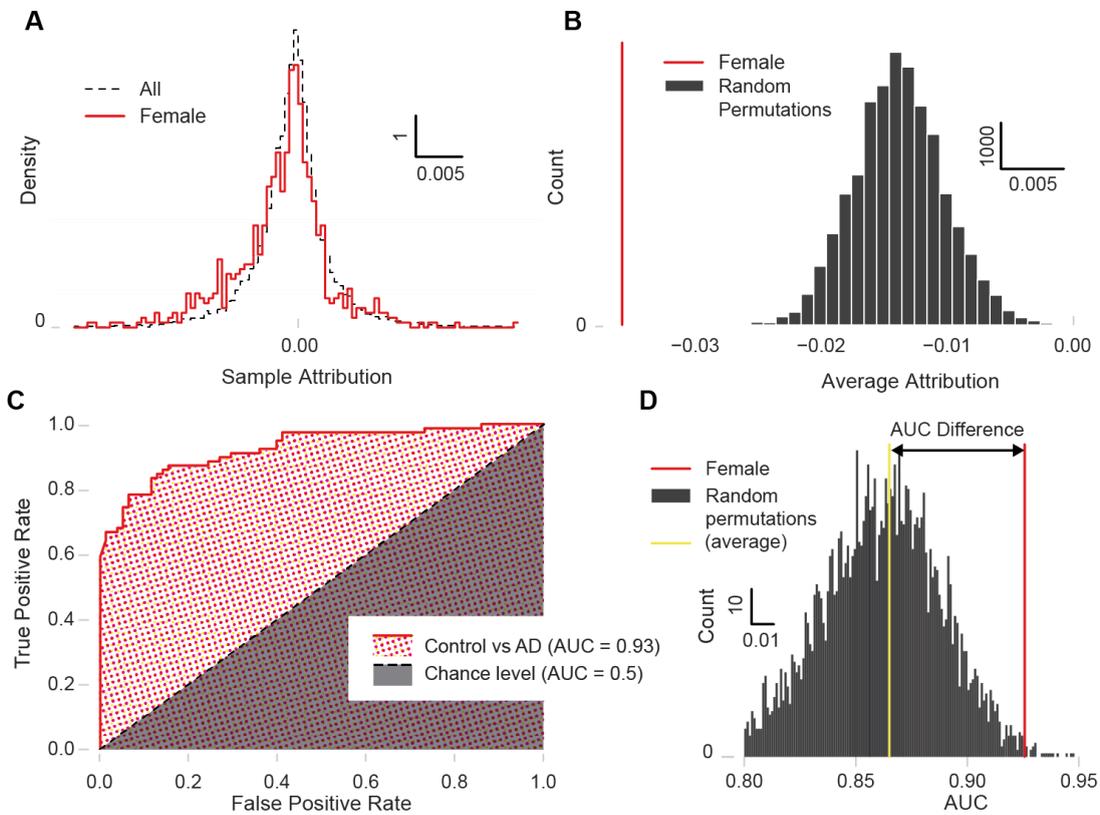

**Figure 2. Statistical Analysis of token attribution and impact on classification. A.** Distribution densities of feature attributions are compared with all tokens to determine group relevance (≤ 0 for control, > 0 for AD). **B.** Using 5000 random subsamples, the feature is checked against the 95th percentile and 5th percentile thresholds. Significant contributions are noted if either condition is met. In this case, the feature *female* contributes to the control group. **C.** The feature is isolated to estimate its classification power using AUC analysis. **D.** The feature's AUC is compared to the mean AUC of the subsamples, with a significant impact indicated if the value is above the 95th percentile or below the 5th percentile.

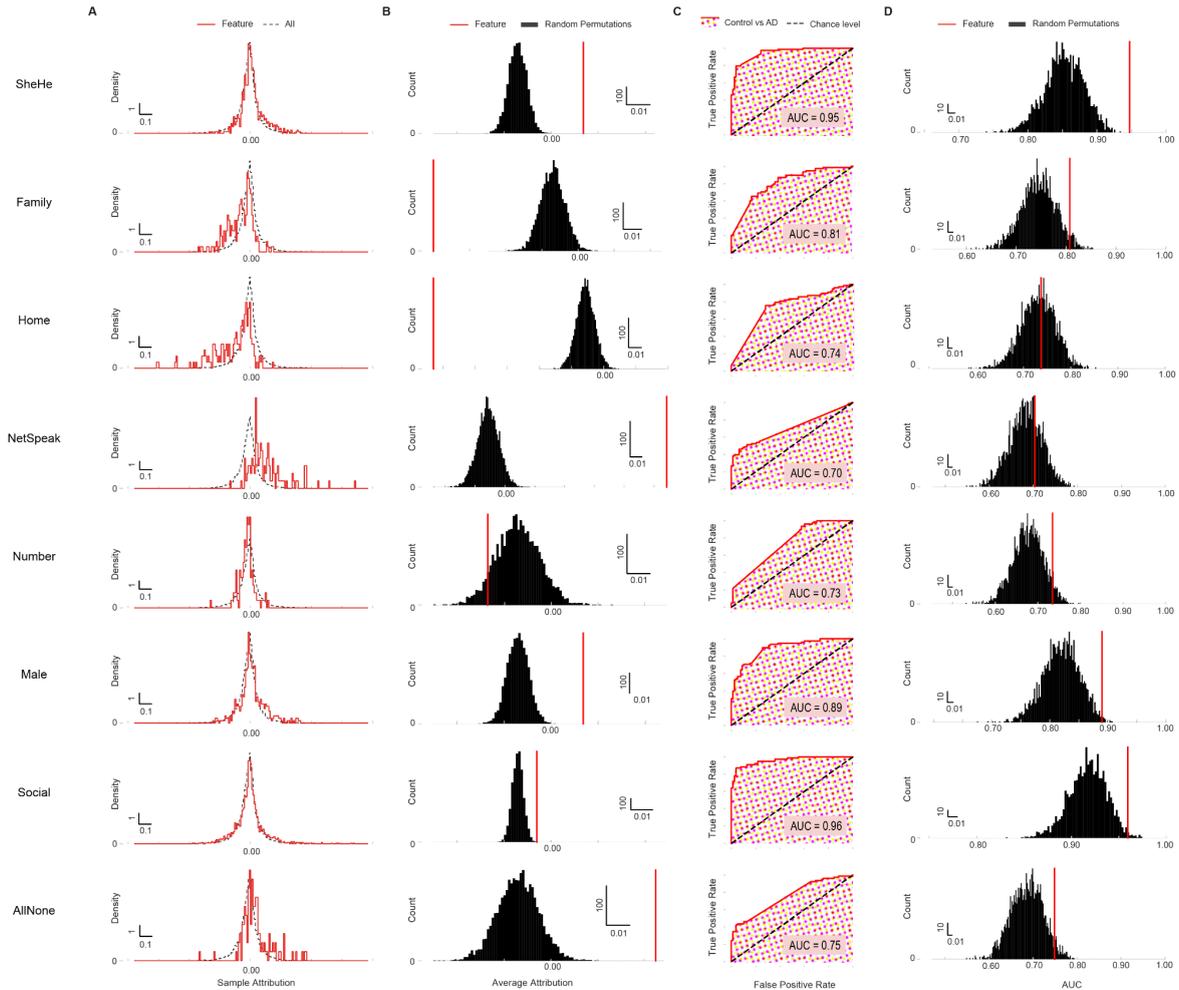

**Figure 3: Statistical analysis of various linguistic features.** The figure illustrates the different possible outcomes related to the significance of linguistic features. (A to D) Same as Figure 2. From top to bottom: the use of *SheHe* (third-person singular pronouns) helps in classifying the AD group (A and B) and has a positive impact on classification (C and D). Words related to *Family* (such as parent, mother, father, baby) contribute to the classification of the control group (A and B) and have a positive impact on classification (C and D). Words associated with *Home* (like home, house, room, bed) contribute to the classification of the control group (A and B) but do not impact the classification (C and D). The same is true for *NetSpeak* (internet language usage like :), u, lol, haha), but it contributes to the AD group (A and B). *Number* words (such as one, two, first, once) do not significantly contribute to any group (A and B) or impact the classification (C and D). References to *Male* (like he, his, him, man) contribute to the classification of the AD group (A and B) and have a positive impact on classification (C and D). *Social* words (generalized words

associated with social processes) contribute to the classification of the control group (A and B) and have a positive impact on classification (C and D). Lastly, *AllNone* words (related to all-or-none thinking, such as all, no, never, always) contribute to the classification of the AD group (A and B) and have a positive impact on classification (C and D).

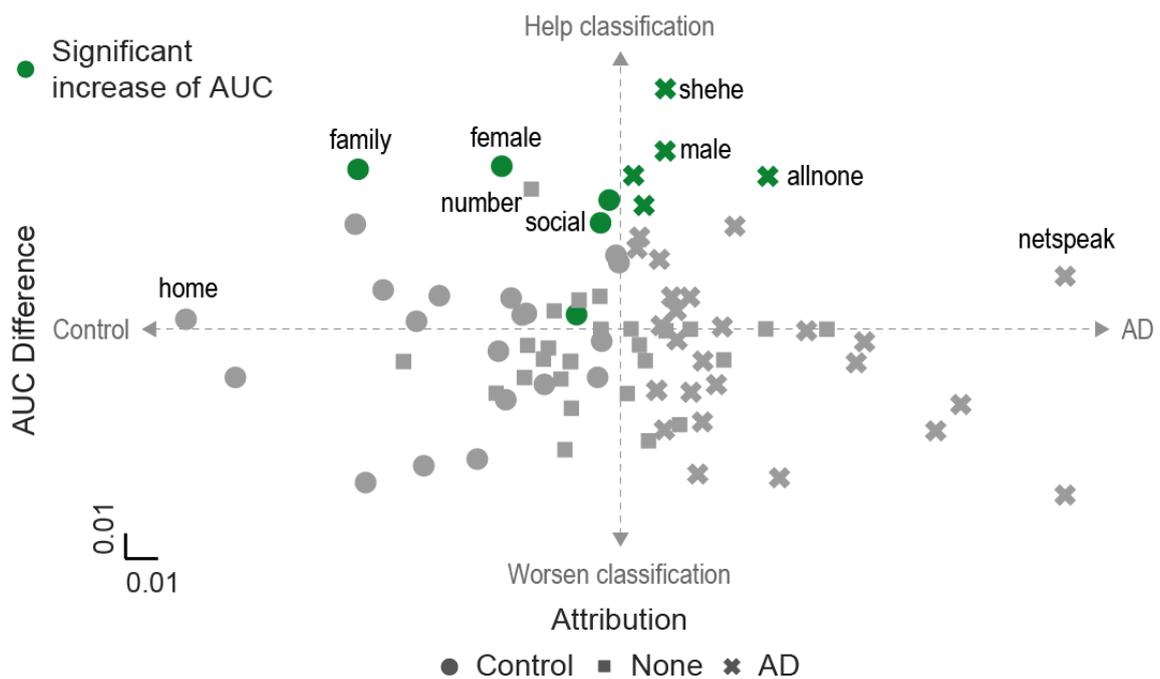

**Figure 4: Relevant linguistic features for group decision and classification.** The figure shows all linguistic features that differ from the mean AUC of random subsamples (y-axis) and the group contribution represented by the feature's attribute value (x-axis). Green indicates features with a positive impact on classification. "X" markers refer to features that contribute to the classification of the AD group, while circles contribute to the classification of the control group.

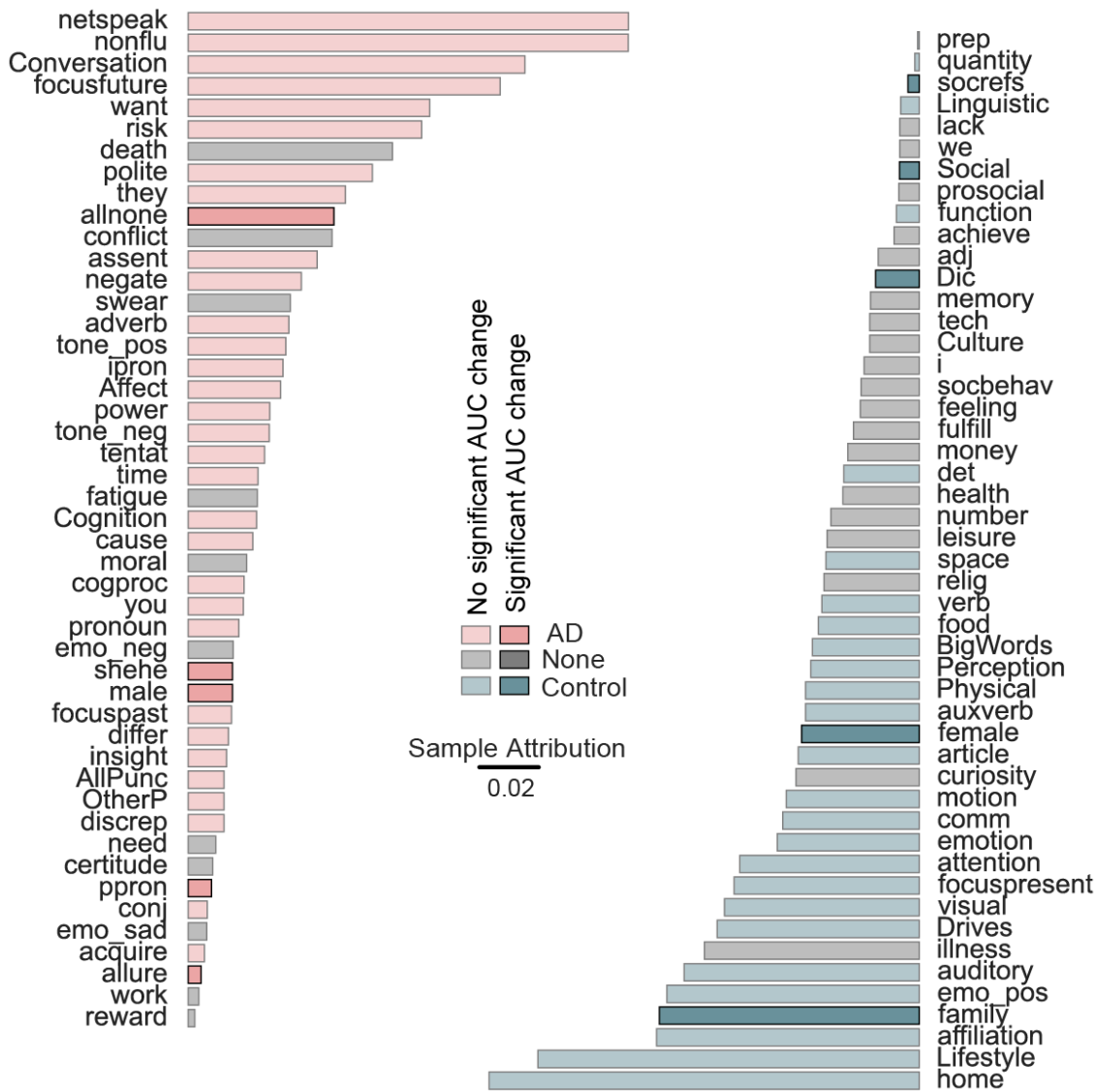

**Figure 5: Bar visualization of relevant linguistic features for group decision and classification.** The figure presents an alternative visualization of the explainability method. The linguistic features that contribute to the decision-making process of the AD group are highlighted in red, while those of the control group are shown in green. Features that do not significantly impact either group are displayed in gray. Features with no significant impact on AUC are displayed with higher transparency. On the left side are the features related to the AD decision that impact AUC: *allnone* (words associated with all-or-none thinking, such as all, no, never, always), *shehe* (third-person singular pronoun), *male* (normative references to males, such as he, his, him, man), *ppron* (personal pronouns), and *allure* (words used in ads and persuasive communications, such as have, like, out, know). On the right side are the features related to the control group decision that impact AUC: *socrefs* (social referents, such as you, we, he, she), *Social* (generalized words associated with social processes), *Dic* (words present in the LIWC dictionary), *female* (normative references to female figures such as she, her, girl, woman), and *family* (words associated with family, such as parent, mother, father, baby).

**Results**

To illustrate the model's functioning, we apply it as a proof-of-concept to an English dataset developed for the ADReSS challenge to detect Alzheimer's disease (AD) through speech analysis. The data consists of audio recordings and transcripts from the Cookie Theft picture description task (Figure 1A). The dataset includes an equal number of subjects in AD and control groups, balanced by age and sex, with 78 participants in each group. Additionally, the dataset contains demographic information such as age and sex and Mini-Mental State Examination (MMSE) scores. The dataset, therefore, provides 156 descriptive narratives (78 AD, 78 control, Figure 1B) with detailed demographic balance, making it suitable for training and validating speech-based AD classification models (Figure 1C).

The method for achieving lexical explainability in text classification tasks consists of three main components: Integrated Gradients (IG) attribution, applying the Linguistic Inquiry and Word Count (LIWC) toolkit, and statistical analysis to identify relevant linguistic features. First, IG is used to compute attribution scores for each token in the input text (Figure 1D). IG quantifies how much each token contributes to the model's final prediction by calculating the gradients of the model's output concerning the input tokens. This provides a detailed map of the words most influential in driving the classification decision, assigning each word a numerical score that reflects its relevance in distinguishing between control and AD subjects. Second, the LIWC toolkit is applied to annotate each token with linguistic categories (Figure 1E). LIWC assigns each word in the input text to one or more predefined linguistic categories, such as pronouns, social references, or function words. Combining IG attribution scores with these linguistic categories, the method identifies which words are essential for classification and reveals the linguistic features (e.g., pronouns, social words) driving the model's decision. Finally, statistical tests are conducted to determine which linguistic features are most relevant for distinguishing between AD and control groups. The attribution scores for tokens belonging to a particular LIWC category are compared with randomly selected tokens to assess whether a specific category is significantly associated with either group (Figure 1F). AUC (Area Under the Curve) analysis measures each feature's classification power, and the significance of each feature's contribution to the model's performance is assessed by comparing these results to random samples (Figures 2 and 3). This approach provides global insights into the linguistic features important for supporting diagnosis and individual-level explanations by highlighting the specific tokens contributing to the model's decision.

The statistical analysis provides two critical insights for the model's explainability: identifying the most relevant linguistic features for the classification and evaluating their impact on classification. These insights can be visualized in different ways. One option is a scatter plot (Figure 4), where each dot represents an explainability feature, plotted according to its attribution value and effect on the AUC. Statistical significance is indicated by variations in the marker's shape and color. Another option is an annotated bar plot (Figure 5), which displays the average attribution of each feature. In this bar plot, statistical relevance is indicated by a color code, highlighting the significance of the attribution value and its corresponding effect on the AUC.

*SLIME attributes provide reliable insights into linguistic features*

Those insights provide a faithful and plausible explanation method. Of the 111 linguistic features analyzed, 64 were relevant for the classification. There are 28 categories for the control group classification and 36 for the AD classification. Five categories improve the accuracy of the model for the classification of the control group: Social referents (*socrefs,* AUC = 0.96)*,* social processes (*Social,* AUC = 0.96)*,* dictionary words (*Dic,* AUC = 0.99)*,* female references (*female,* AUC = 0.93)*,* and family references (*family,* AUC = 0.81); 23 categories contribute to the classification but do not improve the accuracy of the model: linguistic dimension *(Linguistic),* total function words *(function),* determiners *(det),* spatial references *(space),* common verbs *(verb),* food references *(food),* words with 7 letters or longer *(BigWords), Perception, Physical,* auxiliary verbs *(auxverb), article, motion,* communication *(comm), attention,* present focus *(focuspresent), visual, Drives,*

*auditory,* positive emotions *(emo_pos), affiliation, Lifestyle,* and *home*. For the classification of the AD group, five categories improve the accuracy of the model: all-or-none thinking (*allnone,* AUC = 0.75)*,* 3rd person singular pronouns (*shehe,* AUC = 0.95)*,* male references (*male,* AUC = 0.89)*,* personal pronouns (*ppron,* AUC = 0.94)*,* and *allure* (AUC = 0.93); 31 categories contribute to the classification, but do not improve the accuracy of the model: words commonly used on internet (*netspeak),* nonfluencies *(nonflu), Conversation,* future focus *(focusfuture), want, risk, polite,* 3rd person plural pronouns *(they), assent,* negations *(negate), adverb,* positive tone *(tone_pos),* impersonal pronouns *(ipron),* affective words *(Affect), power,* negative tone *(tone_neg),* tentative words *(tentat), time, Cognition, cause,* cognitive processes *(cogproc),* 2nd person pronouns *(you), pronoun,* past focus *(focuspast),* differentiation *(differ), insight,* all punctuation *(AllPunc),* other punctuation (*OtherP),* discrepancy *(discrep),* conjunctions *(conj),* and *acquire*. It's important to note that all linguistic attributes that improve the model's accuracy and contribute to classifying a particular class have accuracy values higher than chance (AUC = 0.5). Thirty-two categories are irrelevant for classification and AUC improvement: *death, conflict, swear, fatigue, moral,* negative emotions *(emo_neg), need, certitude,* sad emotions *(emo_sad), work, reward,* prepositions *(prep), quantity, lack,* 1st person plural pronouns *(we),* prosocial behavior *(prosocial), achieve,* adjectives *(adj), memory,* technology *(tech), Culture,* 1st person singular pronouns *(i),* social behavior *(socbehav), feeling, fulfill, money, health, number, leisure,* religiosity *(relig), curiosity,* and *illness*. The other 15 categories were absent in any tokens: anxiety (*emo_anx),* anger *(emo_anger), friend, politic, ethnicity, wellness, mental, substances, sexual, filler, Period, Comma,* question marks *(QMark),* exclamation points *(Exclam),* and apostrophes *(Apostro)*.

*SLIME attributes offer identity separation capabilities that cannot be captured solely by feature counts*

We then compared the SLIME outcomes with the standard methodology for statistically evaluating how feature counts or frequencies differ between samples from different classes. We assessed the identity separation capability by evaluating the Area Under the Curve (AUC), which measures a model's ability to distinguish between two patterns, with higher AUC values indicating better separation (see Methods). Our findings revealed a strong correlation between AUC values for SLIME attributes and feature counts (Figure 6A, correlation of 0.50). The count-based model identified 10 categories with discriminative power between the AD and control groups. In comparison, the SLIME method identified 64 categories, indicating that SLIME has a greater explanatory and exploratory potential for the condition studied.

Interestingly, only one category — terms related to fulfillment (*fulfill*) — was significant in the count-based model but not in SLIME. This category was less frequent than 89% of the categories identified by SLIME, suggesting that the limited sample size might have influenced the analysis. In contrast, SLIME captured important discriminatory terms associated with socio-psycholinguistic factors, such as male and female references and family-related terms, which the count-based model missed. Both models identified other categories, such as linguistic functions related to punctuation, adverbs, and pronouns, as well as social associations like lifestyle and home-related terms.

The model based on SLIME attributes achieved significantly higher AUC values (Mann-Whitney U test statistic: 7859.5, p-value < 0.001), which aligns with expectations since SLIME attributes are derived from a more advanced model trained using non-linear machine learning techniques. It's crucial to note that SLIME is not designed to be the most accurate or discriminative classifier; instead, its primary purpose is to serve as an explainability method for large language models, which is the focus of this article. Interestingly, the improvement in separation capability was not consistent across all features, leading to a multimodal distribution of changes in AUC (Figure 6B). For most features where the count-based model showed no discriminative power, the SLIME attributes-based model also showed no significant increase in AUC (Figure 6B, gray panel). Among the features that the count-based model did discriminate, some showed no change in AUC, while others exhibited increases of up to 50% (Figure 6B, black panel). A similar pattern was observed for features captured by SLIME (Figure 6B, red panel). Notably, for features that enhanced classification under the SLIME framework, the change in AUC was consistently positive across all features (Figure 6B, blue panel). These findings suggest that the SLIME framework does not indiscriminately enhance explainability but selectively improves the discriminability of features that genuinely impact the classification.

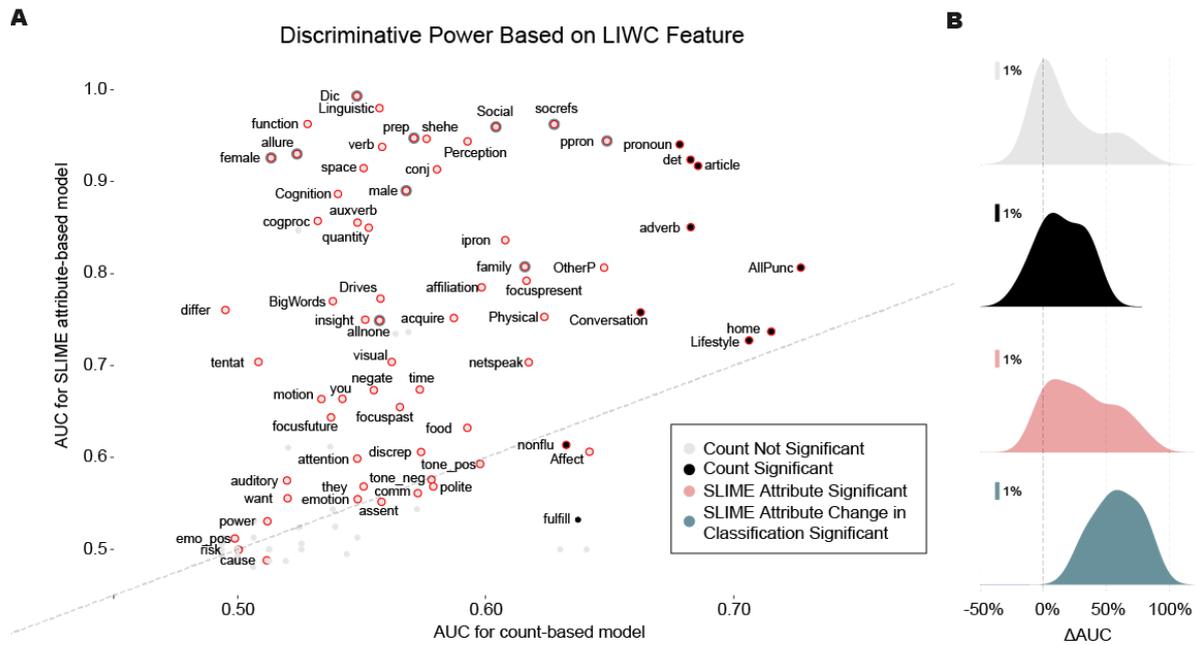

**Figure 6: The SLIME methodology enhances the impact of explainable terms in classification models. A.** Scatter plot illustrating the AUC for classification models based on LIWC features. The x-axis represents the AUC values from LIWC, and the y-axis shows the AUC from SLIME. The data points are color-coded as follows: black dots indicate LIWC features whose counts differ significantly between conditions, gray dots represent features where the counts are not statistically different, red dots mark features where the SLIME attributions are significantly different between conditions, and blue dots highlight features where classification based solely on the SLIME attribution is significantly positive. **B.** Distribution plot displaying the relative difference in AUC between classification models based on SLIME attributions versus LIWC feature counts. Positive values suggest that SLIME attributions offer better class separation than the feature counts. The same feature categories as in panel (A) are used for comparison.

**Discussion**

The proposed methodology advances the field by providing a framework for explaining neural language models (LLMs) in the context of computational linguistics, with a particular focus on Alzheimer's disease (AD) classification. This method leverages Integrated Gradients (IG) for token-level attribution and the Linguistic

Inquiry and Word Count (LIWC) toolkit for linguistic analysis, offering faithful and plausible explanations. In the context of explainability, faithful explanations accurately reflect the model's inner workings (Jacovi & Goldberg, 2020; Rudin, 2019), while plausible explanations make sense to human users, even if they don't fully capture the model's internal processes.

The use of LIWC in this study can be viewed as analogous to tasks like Part-of-Speech (PoS) tagging, where attention mechanisms have been shown to provide faithful explanations (Bibal et al., 2022). Using linguistic categories to annotate the relevant tokens offers interpretable insights into how different linguistic features drive the model's decisions. This aligns with existing literature on AD, which shows that the most relevant linguistic attributes for classification, such as pronoun use and social references, also reflect cognitive changes observed in AD patients (Ahmed, 2013; Almor, 1999; Bucks et al., 2000; Fraser et al., 2016; Lindsay et al., 2021).

The findings demonstrate that the linguistic features identified by the model are computationally significant and clinically relevant. For instance, features like social references and personal pronouns were more prominent in the control group, whereas AD subjects relied more on all-or-none thinking and specific pronouns. It validates the method's ability to produce plausible explanations grounded in observable clinical patterns.

In addition to these insights, this paper also presents a significant technical contribution. The method, implemented in Python, is freely available to other researchers. It offers a generalizable approach based on statistical analysis and linguistic attributes, making it applicable to various binary classification tasks beyond Alzheimer's detection. Notably, the method is not tied to LIWC as the sole source of linguistic annotation; users can input other categories for analysis, enhancing its flexibility for various applications.

This work delivers two key outcomes: first, it provides direct insights into the linguistic markers of AD, confirming existing literature while offering new computational perspectives. Second, it provides a freely available, reproducible method for explaining neural language models, promoting transparency and interpretability across different models and use cases. This approach promotes broader transparency and trust in AI-driven clinical tools by enabling researchers to replicate and adapt the methodology for other domains.

**Acknowledgments**

This work was supported by the Newton Fund and the Royal Society (NAF\R2\202209), CAPES and CNPq. This study used the computational resources of the Núcleo de Processamento de Alto Desempenho (NPAD) of the Federal University of Rio Grande do Norte (UFRN).